# Psychophysical Machine Learning

B.N. Kausik[1]

## Abstract

The Weber Fechner Law of psychophysics observes that human perception is logarithmic in the stimulus. We present an algorithm for incorporating the Weber Fechner law into loss functions for machine learning, and use the algorithm to enhance the performance of deep learning networks.

[1] Unaffiliated independent https://www.linkedin.com/in/bnkausik/; bnkausik@gmail.com

## Introduction

The origins of psychophysics date back to the 1800s when Gustav Weber and Ernst Fechner studied the relationship between physical stimulus and human perception, (Fechner 1866). Originally formulated as two separate observations - Weber's Law and Fechner's law, the combined Weber-Fechner Law observes that human perception varies as the logarithm of the physical signal or stimulus. Basic sensory functions such as vision, weight, and sound follow the law, e.g., we measure sound level on the logarithmic dB scale. Cognitive functions such as distinguishing between numbers also follow the law, (Moyer and Landauer 1967) (Mackay 1963), (Staddon 1978), and (Longo and Lourenco 2007).

Since machine learning is typically concerned with learning human concepts, it is reasonable to ask whether the Weber Fechner logarithmic law can improve learning algorithms. Previously, (Kausik 2022) showed that a logarithmic power series of basic predictors can enhance learning. In this paper, we present a general methodology for incorporating the Weber Fechner Law into the loss function of any learning algorithm. Our experiments show that such loss functions can enhance the performance of deep learning neural networks.

Of related interest is the augmentation of training sample labels with the human psychophysical reactions observed during classification, e.g. (Dulay et al, 2022).

## Preliminaries

Let $I = [0, 1]$ be the unit interval, and $I^n$ be the unit ball in $n$ dimensions. We consider the learning of multiclass concepts $C: I^n \rightarrow \{1, 2, \ldots m\}$ that map inputs $x \in I^n$ to one of $m$ classes. A learning algorithm takes a set of labeled training samples $\{(x, l)\}$ and outputs a function $F: I^n \rightarrow \{1, 2, \ldots m\}$, where $F(x) = \{p_1, p_2, \ldots, p_m\}$ is a probability distribution over $\{1, 2, \ldots m\}$, i.e. $\sum_{i=1}^{m} p_i = 1$. Specifically, $p_i$ is the probability that $C(x) = i$ as predicted by function $F$, and $argmax(p_1, p_2 \ldots p_m)$ is the predicted class for input $x$.

We consider learning algorithms that construct $F$ to minimize a chosen loss function $L$, which measures the accuracy in predicting $C$. The loss function is typically a continuous function, to aid in iterative gradient descent methods to construct the function $F$. For multiclass learning, the loss function is typically chosen to be the Sparse Categorical Cross Entropy. On a set of samples $T = \{(x, l)\}$,

$$L_e(T, F) = - \sum_{(x,l) \epsilon T} log(F(x)_l).$$

The Weber Fechner Law states that $P = Klog(S)$, where $P$ is the perception and $S$ is the stimulus. In the case of physical concepts such as sound level and weight, it is clear that the stimulus is directly the physical signal. In the case of cognitive concepts, we interpret the stimulus $S(x)$ associated with input $x$ as a measure of the deviation from some unknown and hypothetical ideal of the concept. Likewise, we interpret the perception as

$$P(x) = p_i : i = C(x)$$

as assigned by a human observer of input $x$.

We now consider the sensitivity of $S(x)$ to multiplicative random noise on input $x$. Specifically, for input $x$ and random noise $\vartheta \epsilon R^n$ of strength $||\vartheta|| = \delta$ and zero mean, consider the perturbation $x + x \star \vartheta$, where the $\star$ denotes element-wise multiplication. Normalizing, we set

$$x' = (x + x \star \vartheta)/||x + x \star \vartheta||.$$

Multiplicative random noise is proportional to the signal at all points, thereby distorting weak and strong portions of the signal proportionally. In contrast, additive random noise can swamp weak but important features of the signal.

**Assumption:** Multiplicative random noise diminishes the cognitive signal in the normalized stimulus proportionally in that

$$E[S(x')] = (1 - \delta)S(x), \quad (1)$$

where $E$ denotes the expectation over the noise variable.

Per the Weber Fechner Law, we have

$$P(x) = Klog(S(x)) \text{ and } P(x') = Klog(S(x')).$$

Rearranging, we get

$$P(x') = P(x) - K(log(S(x)) - log(S(x')))$$
$$= P(x) - Klog(S(x)/S(x'))$$
$$= P(x) + Klog(S(x')/S(x))$$

Taking the expected value on both sides of the above and invoking Equation (1), we get

$$E[P(x')] = P(x) + Klog(1 - \delta). \quad (2)$$

## The Modified Loss Function

We now introduce the Weber Fechner loss function, which estimates the deviation of $F$ from the Weber Fechner law. On a set of samples $T = \{(x, l)\}$,

$$L_w(T, F) = - \sum_{(x,l)\in T} \Big(F(x)_l + K\log(1 - \delta)\Big) \log\Big(E\big[F(x')_l\big]\Big), \quad (3)$$

where $x' = x + \vartheta$ and $E$ denotes the expectation. Since we do not know $K$ a priori, we estimate it empirically as follows

$$K = \frac{1}{\log((1-\delta))} \sum_{(x,l)\in T} E(F(x')_l) - F(x)_l \quad (4)$$

It is important to note that minimizing the Weber Fechner loss function imposes conformance with the law, but does not enforce fit with the labels in the training samples. To enforce the latter, we use a weighted sum of the Weber Fechner loss and the Entropic loss in a composite loss function

$$L = wL_w(T,F) + (1 - w)L_e(T,f).$$

Intuitively, minimizing $L_w$ imposes the Weber Fechner Law to enhance generalization, while minimizing the cross entropy $L_e$ improves the fit on the training samples. Algorithm 1 shows the steps involved.

---
**Algorithm 1: Modified Loss Function**
---

**Input:** $T = \{(x, l)\}$, $F$
**Parameter:** $w$, $\delta$, $m$

**Compute** $K\log(1 - \delta) = \sum_{(x,l)\in T} E(F(x')_l) - F(x)_l \quad (5)$

and $L_w = -\sum_{(x,l)\in T} \Big(F(x)_l + K\log((1 - \delta))\Big)\log\Big(E\big[F(x')_l\big]\Big) \quad (6)$

where for each $x$, the expectation is taken across $x' = (x + \vartheta)/\|x + \vartheta\|$, for $m$ samples of random noise $\vartheta$.

**Compute** $L_e = - \sum_{(x,l)\in T} \log(F(x)_l)$.

**return** $wL_w + (1 - w)L_e$

---

## Experimental Results

We implemented the modified loss algorithm in the open source Tensorflow system (www.tensorflow.org) on a Macbook Pro laptop running the Catalina operating system. We tested the system on the MNIST data set (Lecunn, Cortes and Burgess, 2022) for cursive characters which comprises 60,000 training samples of handwritten digits and 10,000 test samples. Each sample is a 28x28 image, with an associated label {0,1,...9}. See Fig.1.

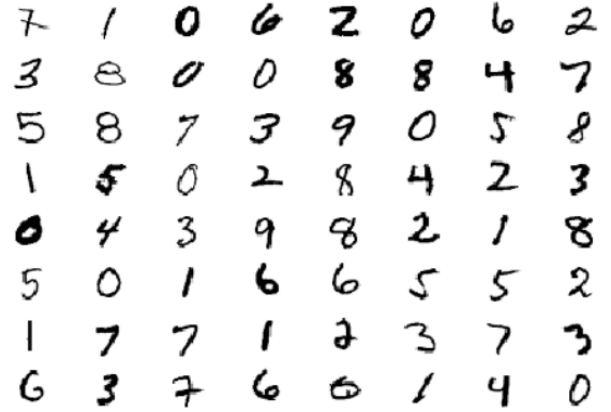

Fig.1: Sample MNIST digits

To start, we chose the off-the-shelf linear network specified in the Tensorflow online MNIST tutorial as in Fig.2.

```
Layer (type)                 Output Shape              Param #
=================================================================
flatten (Flatten)            (None, 784)               0
dense (Dense)                (None, 128)               100480
dropout (Dropout)            (None, 128)               0
dense_1 (Dense)              (None, 10)                1290
=================================================================
Total params: 101,770
Trainable params: 101,770
Non-trainable params: 0
```

Fig.2: Linear Neural Net

For Algorithm 1, we chose Gaussian noise of zero mean and variance 0.05, implying $\delta \approx 0.05$ or 5%. The weight parameter $w$ ranged across {0.0, 0.1, 0.2,...0.5}. At each value of $w$, we ran 10 iterations of training and testing of the network. The training phase was 20 epochs, with the number of noise variations $m = 1$ per epoch per sample. In a high-dimensional input space such as MNIST, $n = 28x28 = 784$, and each noise vector draws 784 independent random noise scalars, aiding convergence. Across iterations, we used an exponential moving average of the estimate for $K\log(1 - \delta)$, with a decay factor of 0.99. After training, we evaluated the training network on the 10,000 test samples.

Fig. 3a shows the results of our experiments on the MNIST dataset. Each point in the figure is the average of the 10 underlying iterations. Setting $w = 0$ essentially uses sparse categorical cross entropy loss, while increasing $w$ apportions increasing weight to Weber Fechner loss. The error bars denote $\frac{\sigma}{\sqrt{n}}$, where $n = 10$ is the number of iterations, and σ is the observed standard deviation. It is clear that setting the $w = 0.3$ delivers the best results in this case.

Many prior works, e.g., (Bishop, 1995), (Ciresan et al, 2012), report that including distorted training samples in the training phase can improve performance. To that end, we included the perturbations $x'$ of Algorithm 1 in the training set at $w = 0$ but found no measurable difference in performance with or without. This suggests the gains are derived from the modified loss function rather than the noise perturbations.

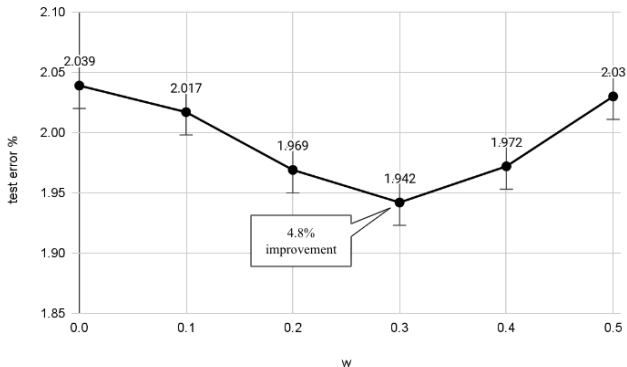

Fig. 3a: MNIST test error vs Weber Fecher loss weight (linear network)

Fig. 3b shows the training error for the linear network on the MNIST dataset. The training error was measured on both the original and the noisy samples.

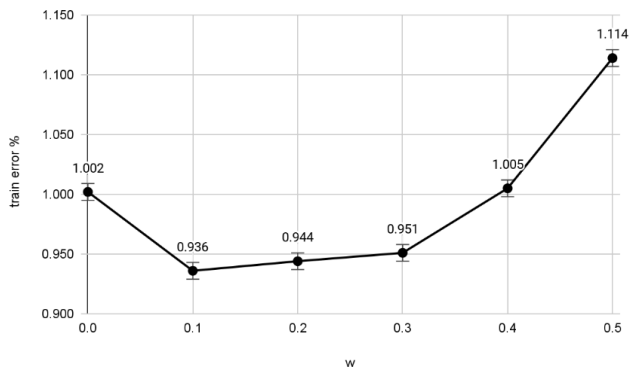

Fig. 3b: MNIST train error vs Weber Fecher loss weight (linear network)

Fig. 3c shows the total training loss for the linear network on the MNIST dataset. While the test error and train error of Figs. 3a and 3b roughly track each other, the total loss does not. This is understandable, since the Weber Fechner loss and Entropic loss are not directly comparable. The implication is that the optimal value of $w$ must be determined by other means, e.g. via a control subset of the training set for the purpose of evaluating $w$, or increasing the number of iterations and epochs for sharper convergence in the training error of Fig. 3b.

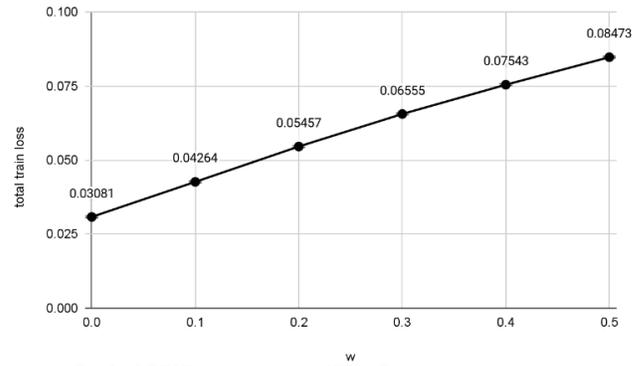

Fig. 3c: MNIST total train loss vs Weber Fecher weight (linear network)

Our modified loss function is similar in spirit to regularization methods, e.g. (Bishop 1995), (Srivastava et al., 2014). Regularization seeks to simplify the learnt function by introducing a direct complexity measure in the loss function, or indirectly via noise, dropout or other perturbations in the training samples. In contrast, our method seeks to enforce the Weber Fechner Law, by introducing a term in the loss function that is evaluated on perturbations of the training samples. In short, regularization enforces Occam's Razor while our method enforces the Weber Fechner Law. Both methods can be used in combination as we have done in our experiments, combining dropout regularization in the neural network model with Weber Fechner loss calculation.

We also tested an off-the-shelf convolutional neural network for the MNIST dataset (keras.io/examples/vision/mnist_convnet/) as shown in the following.

```
Layer (type)                 Output Shape              Param #   
=================================================================
conv2d (Conv2D)              (None, 26, 26, 32)        320       
max_pooling2d (MaxPooling2D  (None, 13, 13, 32)        0         
)                                                                
conv2d_1 (Conv2D)            (None, 11, 11, 64)        18496     
max_pooling2d_1 (MaxPooling  (None, 5, 5, 64)          0         
2D)                                                              
flatten_1 (Flatten)          (None, 1600)              0         
dropout_1 (Dropout)          (None, 1600)              0         
dense_2 (Dense)              (None, 10)                16010     
=================================================================
Total params: 34,826
Trainable params: 34,826
Non-trainable params: 0
```

Fig.4: Convolutional Neural Net

Figs. 5a and 5b show the test error and train error respectively for the convolutional network on the MNIST dataset. The training error was measured on both the original and the noisy samples.

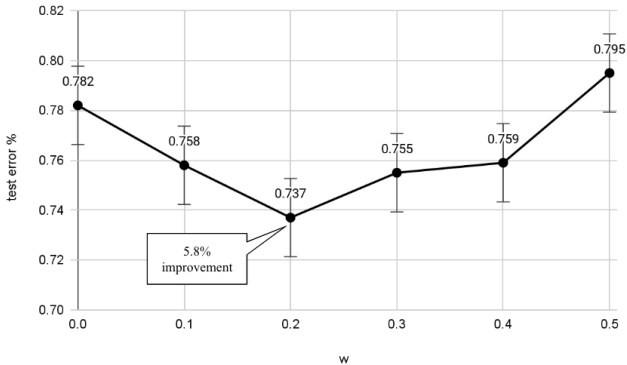
Fig. 5a: MNIST test error vs Weber Fechner loss weight (convolutional network)

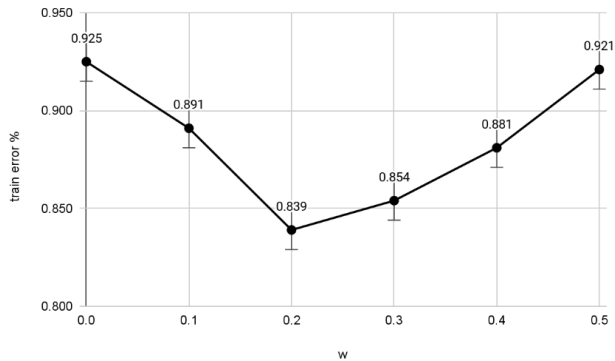
Fig. 5b: MNIST train error vs Weber Fechner loss weight (convolutional network)

We also ran experiments on the KMNIST dataset of handwritten Japanese characters, (Clanuwat et al., 2018). As with MNIST, the training and test sets comprise 60,000 and 10,000 labeled samples respectively, and each sample is a 28x28 image, with an associated label {0,1,...9}. See Fig. 6.

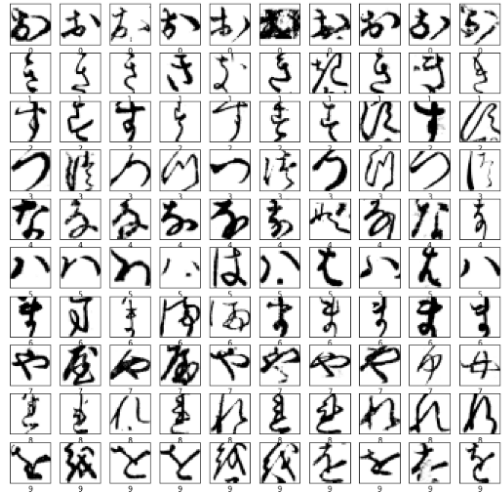
Fig.6: Sample KMNIST characters

Figs. 7a and 7b show the test error and train error of the linear network on the KMNIST dataset. The linear network performed poorly to start with, and improved a bit with the modified loss function.

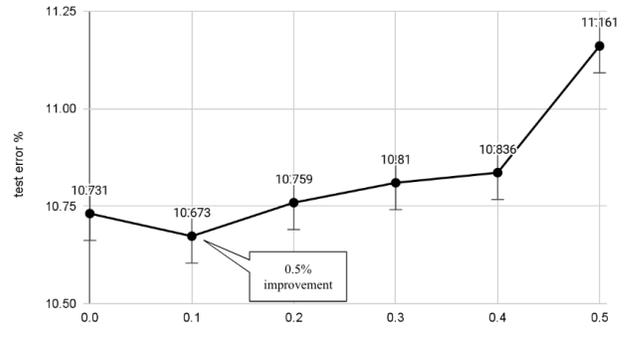
Fig. 7a: KMNIST test error vs Weber Fecher loss weight (linear network)

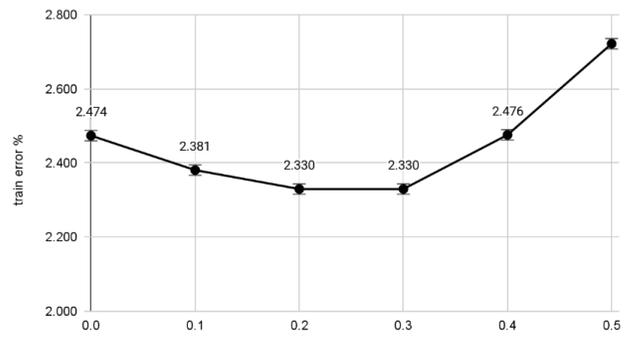
Fig. 7b: KMNIST train error vs Weber Fecher loss weight (linear network)

Lastly, Figs. 8a and 8b show the test and train error of the convolution network on the KMNIST dataset.

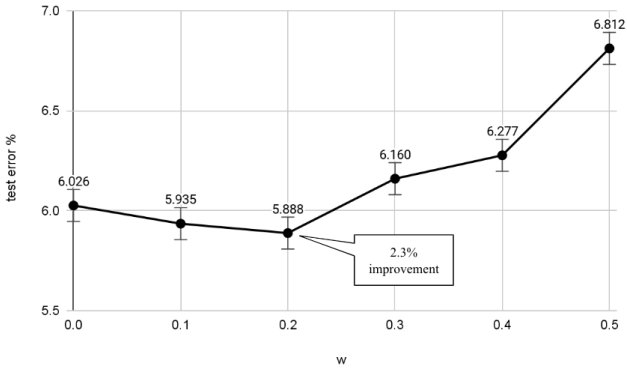

Fig. 8a: KMNIST test error vs Weber Fecher loss weight (convolutional network)

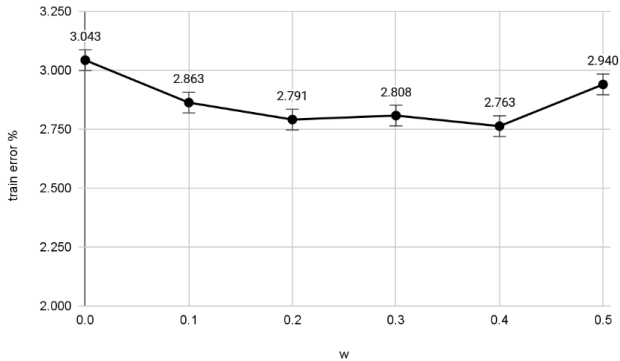

Fig. 8b: KMNIST train error vs Weber Fecher loss weight (linear network)

The convolutional network performs modestly on KMNIST to start with, and enjoys modest gains with the modified loss function.

## Variations

In this section, we consider alternatives to multiplicative Gaussian noise for calculating the Weber Fechner loss.

Firstly, for sample $x$, we use randomly chosen alternate training samples $x'$ of the same label. In this case, δ is the average of $|S(x) - S(x')|$ across training samples with the same label. Since Equation (5) iteratively estimates $Klog(1 - δ)$, we do not need to estimate δ separately. However, since Equation (1) no longer applies, we flip the sign of Equation (6) in Algorithm 1 when $F(x')_l > F(x)_l$.

Fig. 9a shows the test error for the linear network on MNIST using random alternate samples for the modified loss calculation.

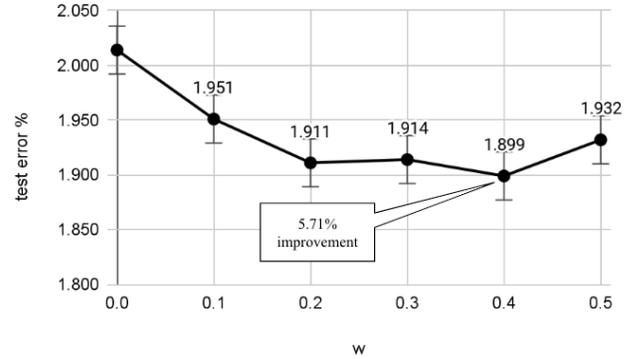

Fig. 9a: MNIST test error vs w using alternates (linear network)

Fig. 9b shows the test error for the convolutional network on MNIST using random alternate samples for the modified loss calculation.

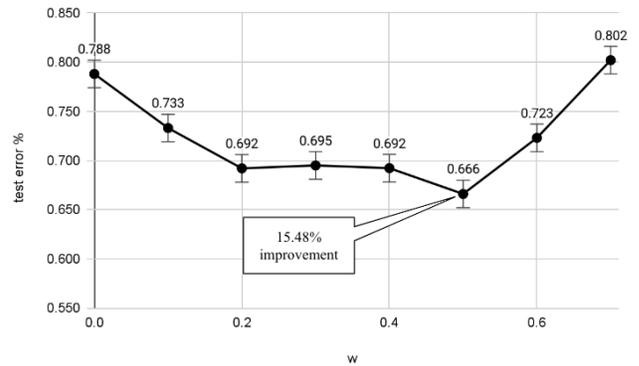

Fig. 9b: MNIST test error vs w using alternates (convolutional network)

Next, we consider strong binary multiplicative noise. Specifically, $x'$ randomly sets 25% of the pixels of $x$ to zero and normalized such that $||x'|| = 1$. Fig. 9a shows the test error for the linear network on MNIST, delivering ~2X the improvement of Fig. 3a. Recall that with weak Gaussian noise, training on the perturbed samples using the unmodified cross entropy loss delivered no measurable improvement. In contrast, strong binary noise delivered ~5% improvement when training on the perturbed samples using the unmodified cross entropy loss function.

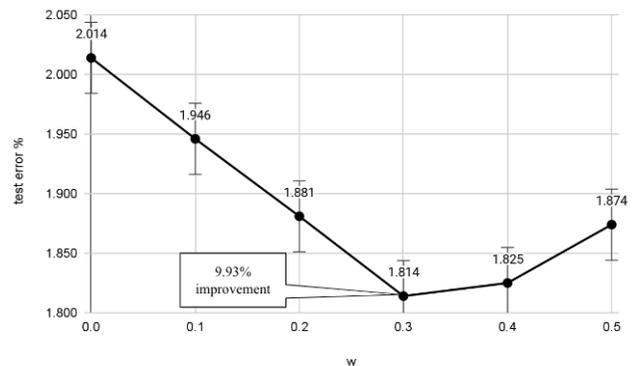

Fig. 9a: MNIST test error vs w using 25% binary noise (linear network)

Strong binary noise was ineffective on convolutional networks in our experiments.

## Summary

The Weber-Fechner Law observes that human perception varies as the logarithm of the physical signal or stimulus. We presented a general methodology for incorporating the Weber Fechner Law into the loss function of any learning algorithm. Our experiments show that such loss functions can enhance the performance of deep learning neural networks, thereby encouraging further research on including the Weber Fechner Law in machine learning.

## Acknowledgements

Thanks to Jay Jawahar, Steven Johnson and Prasad Tadepalli for comments and suggestions.